\documentclass[10pt,twocolumn,letterpaper]{article} 

\usepackage{avss}
\usepackage{times}
\usepackage{epsfig}
\usepackage{graphicx}
\usepackage{amsmath}
\usepackage{amssymb}

\usepackage{fancyhdr}
\usepackage[usestackEOL]{stackengine}
\usepackage{enumerate}
\usepackage{multirow}
\usepackage{array}
\usepackage{color}
\usepackage{subcaption}
\usepackage{url}
\usepackage[linesnumbered,ruled,vlined]{algorithm2e}
\usepackage{tabularx}

\renewcommand{\ieeecopyright}{978-1-5386-2939-0/17/\$31.00 \copyright2017 IEEE}

\avssfinalcopy 

\def\avssPaperID{11} 

\ifavssfinal\fi
\begin{document}
\title{Using Deep Networks for Drone Detection}

\author{Cemal Aker, Sinan Kalkan\\
KOVAN Research Lab.\\
Computer Engineering, Middle East Technical University\\
Ankara, Turkey\\
{\tt\small \{cemal, skalkan\}@ceng.metu.edu.tr}
}
\maketitle

\begin{abstract}
Drone detection is the problem of finding the smallest rectangle that encloses the drone(s) in a video sequence. In this study, we propose a solution using an end-to-end object detection model based on convolutional neural networks. To solve the scarce data problem for training the network, we propose an algorithm for creating an extensive artificial dataset by combining background-subtracted real images. With this approach, we can achieve precision and recall values both of which are high at the same time.
\end{abstract}

\section{Introduction}
Drone, as a general definition, is the name coined for the unmanned vehicles. However, in this paper the term will refer to a specific type, namely unmanned aerial vehicles (UAV). With the rapid development in the field of unmanned vehicles and technology used to construct them, the number of drones manufactured for military, commercial or recreational purposes increases sharply with each passing day. This situation poses crucial privacy and security threats when cameras or weapons are attached to the drones. Hence, detecting the position and attributes, like speed and direction, of drones before an undesirable event, has become very crucial.
\par
Unpredictable computer controlled movements, speed and maneuver abilities of drones, their resemblance to birds in appearance when observed from a distance make it challenging to detect, identify and correctly localize them. In order to solve this problem, one can think of various types of sensors to perceive the presence of a drone in the environment. These may include global positioning systems, radio waves, infrared, and audible sound or ultrasound signals. However, it has been reported that they have many limitations for this problem, and suggested that computer vision techniques be used \cite{gokcce2015vision}. Although deep learning methods have been shown to be very powerful in computer vision tasks, the studies to detect UAVs have not taken advantage of it by placing deep learning methods at the core of the approach. To this end, this study is the first to evaluate the success of convolutional neural networks (CNN) as a standalone approach on drone detection.
\par
In this study we have used an end-to-end object detection method based on CNNs to predict the location of the drone in the video frames. In order to be able to train the 
network, we created an artificial dataset by combining real drone and bird images with different background videos. The results show that the variance and the scale of the dataset make it possible to perform well on drone detection problem. With this method, we have participated in the Drone-vs-Bird Detection Challenge\footnote{\url{https://wosdetc.wordpress.com/challenge}} organized within the International Workshop on Small-drone
Surveillance, Detection and Counteraction Techniques, and our trained network ranked third in terms of lowest prediction penalty described in Section~\ref{exp}.

\begin{figure}[t]
	\includegraphics[width=\columnwidth]{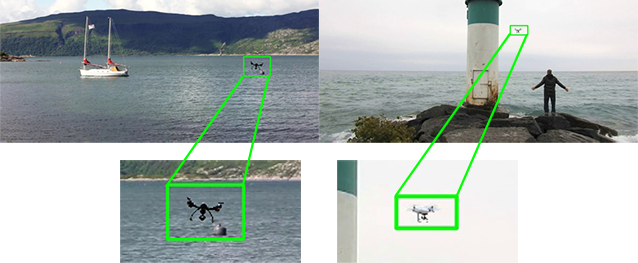}
    \caption{Detection samples from the created dataset where the green rectangles show the bounding boxes of the drones.}
\end{figure}
\section{Related Work}
In this section, we review the related studies in two parts. 

\subsection{Object Detection Methods with Computer Vision}
The task of object detection is to decide whether there are any predefined objects in a given image or not, and report the locations and dimensions of the smallest rectangles that bind them if they exist. Early attempts for this task involves the representations of objects using handcrafted features whereas the state of the art techniques utilizes deep learning.

\textbf{Detection with Handcrafted Features:}
The most successful approaches using handcrafted features require bag of visual words (BoVW) \cite{sivic2003video} representations of the objects with the help of local feature descriptors such as scale invariant feature transform (SIFT) \cite{lowe1999object}, speeded-up robust features (SURF) \cite{bay2008speeded}, and histogram of oriented gradients (HOG) \cite{dalal2005histograms}. After training a discriminative machine learning model, \eg, support vector machines (SVM) \cite{cortes1995support}, with such representations, the images are scanned for occurrence of learned objects with the sliding window technique generally. These methods have two crucial drawbacks. The first one is that the features have to be crafted well for the problem domain to highlight and describe the important information in the image. The second one is the computational burden of the exhaustive search done by the sliding window technique.

\textbf{Detection with Deep Networks:}
With the remarkable achievements of the deep learning methods in the image classification tasks, similar approaches have started to be used for attacking the object detection problem. These techniques can be divided into two simple categories; region proposal based and single shot methods. The approaches in the first category differs from the traditional methods by using features learned from data with CNNs and selective search or region proposal networks to decrease the number of possible regions \cite{girshick2015fast, girshick2014rich, ren2015faster}. In the single shot approach, the aim is to compute bounding boxes of the objects in the image directly instead of dealing with regions in the image. A method for this is extracting multi-scale features using CNNs and combining them to predict bounding boxes \cite{he2014spatial, liu2016ssd}. Another one, named YOLO, divides the final feature map into a 2D grid and predicts a bounding box using each grid cell \cite{redmon2016you}.

\subsection{UAV Detection Methods with Computer Vision}
Although the problem of detecting UAVs is not a well studied subject, there are some attempts to mention. Mejias \etal utilized morphological pre-processing and Hidden Markov Model filters to detect and track micro unmanned planes \cite{mejias2010vision}. G\"{o}k\c{c}e \etal used cascaded boosted classifiers along with some local feature descriptors \cite{gokcce2015vision}. In addition to this pure spatial information based methods, spatio-temporal approaches exist. Rozantsev \etal propose a method that first creates spatio-temporal cubes using sliding window method at different scales, applies motion compensation to stabilize spatio-temporal cubes, and finally utilizes boosted tree and CNN based regressors for bounding box detection \cite{EPFL-ARTICLE-218330}.
\section{Method} \label{method}
Our solution is based on a single shot object detection model, YOLOv2 \cite{DBLP:journals/corr/RedmonF16}, which is the follow-up study of YOLO. We adapt and fine-tune this model to detect objects of two classes (\ie, drone and bird). Although the problem is detecting drones in the scene, we have included the bird class so that the network can learn robust features to distinguish them too. In order to achieve high accuracy with such deep models, one needs a large scale dataset that includes many scenarios of the problem, to get better generalization. To this end, we created an artificial dataset including real drones, real birds and real backgrounds. The following paragraphs first describe the approach in YOLOv2, the dataset creation approach, training and testing details.
\par
\subsection{The deep network} 
\begin{figure*}[t]
	\includegraphics[width=\textwidth]{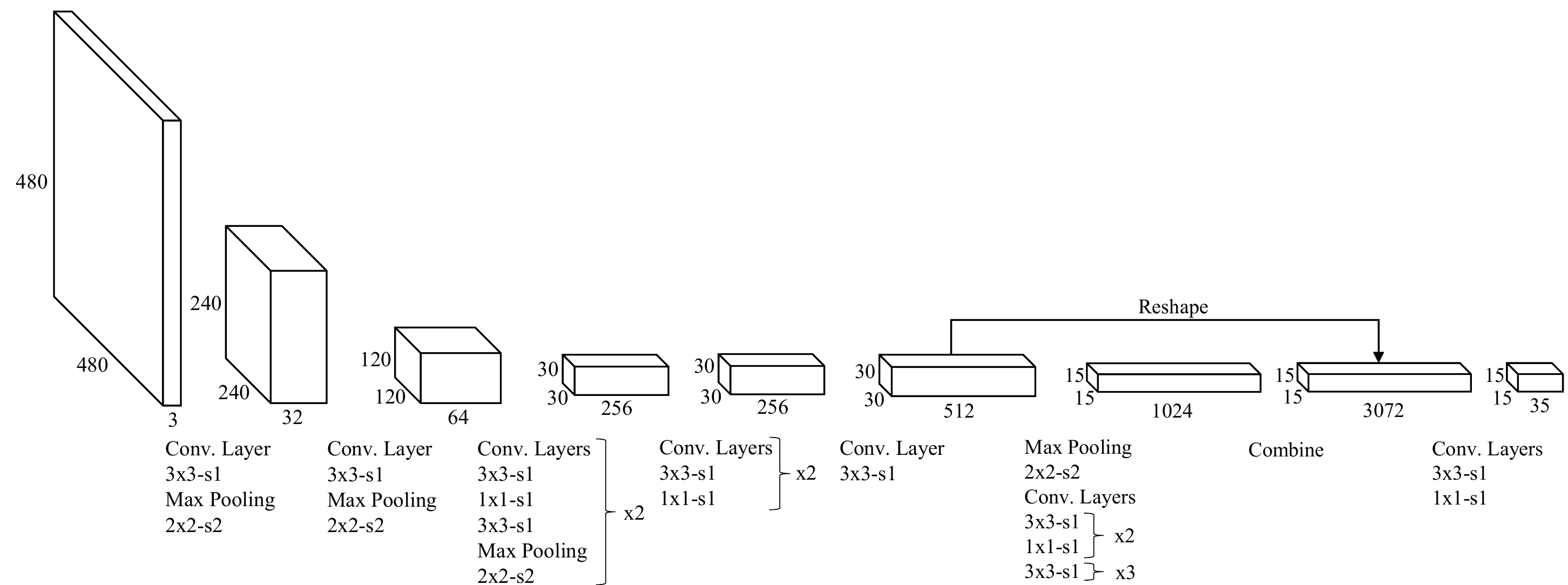}
    \caption{Our adaptation of the YOLOv2 network. All layers are fine-tuned with the dataset collected in the paper.}
    \label{yolomodel}
\end{figure*}

YOLOv2 tries to devise an end-to-end regression solution to the object detection problem. Former layers of the fully convolutional architecture that can be seen in Figure \ref{yolomodel} are trained to extract high level features. Then the two highest level features are combined to get the final feature map of the image. Then it is divided into an $S \times S$ grid where the duty of each grid cell is predicting  bounding boxes of the form $(x,y,w,h,c)$. In this output, $x$ and $y$ are the coordinates of the centers of the boxes with respect to the grid cell, $w$ and $h$ are the width and height in proportion to the whole image, and $c$ is the confidence that an object is in the bounding box. The final task of a grid cell is computing conditional class probabilities given the probability that the corresponding bounding boxes have objects in them. While predicting those bounding boxes, the model utilizes some prior information computed by K-means clustering on width and heights of ground truth bounding boxes. The final output size for a grid cell is:
\[Output\:Size = (N_{cls}+N_{coord}+1)\times N_{anc},\]
where $N_{cls}$ is the number of classes, $N_{coord}$ is the number of coordinates, $N_{anc}$ is the number of anchor bounding boxes used as prior knowledge and the 1 in the parenthesis is for the confidence value. In our approach, grid size is set to 15, number of classes is two, number of coordinates is four and number of anchor boxes is five. Hence, the final output is of the shape $15\times 15 \times 35$.

\subsection{Dataset Preparation}
\begin{figure}
	\centering
		\includegraphics[width=\columnwidth]{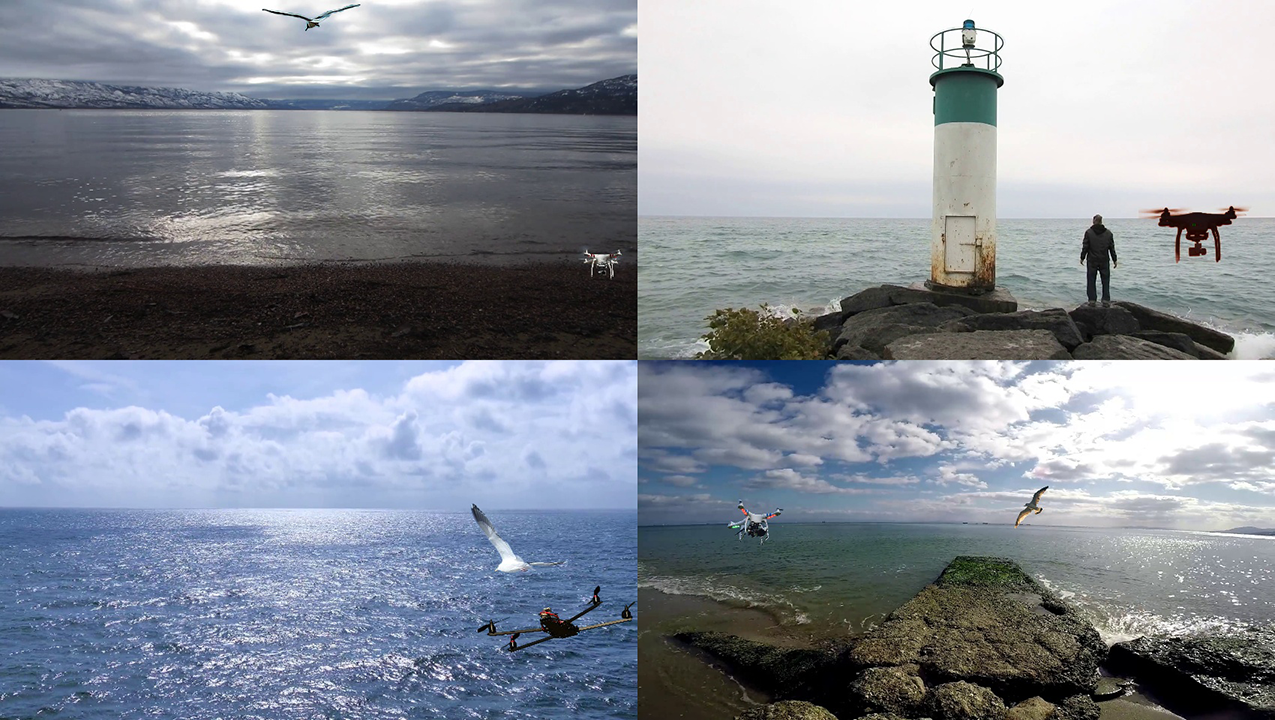}
    	\caption{Samples from the artificial dataset which represent various scenarios with different backgrounds and bird inclusion. Although the dataset includes very small objects, the bigger ones have been chosen for better visibility. (best viewed in color).}
        \label{samples}
\end{figure}

Having mentioned the model details,  we can now come to the most important part of the study which is dataset preparation. Since drone flights have limitations due to inadequate battery technology, weather conditions and legislative regulations, there is no publicly available large scale dataset for training deep networks. However, our approach requires immense amount of data to learn useful features. One possible solution to this is creating an artificial dataset. For this end, we have collected public domain pictures of drones and birds, and videos of coastal areas. After subtracting the background of drones and birds, they are randomly placed on the frames of the videos. The overall process is summarized in Algorithm \ref{dataset}. The details of the dataset can be found in the Table \ref{datasettable}. As can easily be seen, the dataset needs a huge storage size when all of the configurations are used. Hence, we eliminated some portion of the configurations with probability 
\[p=1-\frac{Max.\:allowed\:size}{Total\:size\:for\:all\:configurations},\]
to reduce the size of the dataset to reasonable amounts. Samples drawn from the resulting dataset can be seen in the Figure \ref{samples}. These samples show that although they are created artificially, they look like real images of flying drones and birds.

\begin{algorithm}[!h]
\caption{The algorithm for preparing the dataset.}
\label{dataset}
\SetAlgoLined
\DontPrintSemicolon

$S \gets \text{predefined size intervals}$\;
$D \gets \text{foregrounds of drone images}$\;
$B \gets \text{foregrounds of bird images}$\;
$V \gets \text{background videos}$\;
$R \gets \text{\# of rows that the image will be divided into}$\;
$C \gets \text{\# of columns that the image will be divided into}$\;
$G \gets R\times C \text{ grid}$\;

\ForEach{$(d,g,s,v) \in D\times G \times S \times V$}
{
  ignore this configuration with probability $p=1-\frac{Max.\:allowed\:size}{Total\:size\:for\:all\:configurations}$, and continue\;

  draw a random position $p_0$ in $g$\;
  draw a random size $s_0$ for smaller edge of the drone from $s$\;
  draw a random frame $f_0$ from $v$\;
  resize $d$ with respect to $s_0$\;
  overlay $f_0$ with $d$ in position $p_0$\;

  draw $(p_1,s_1,f_1)$ in the same way\;
  draw a random bird $b_0$ from $B$\;
  draw $(p_{b,0},s_{b,0})$ for bird where $s_{b,0}$ is drawn from smaller half of $S$\;            
  resize $d$ with respect to $s_1$\;
  overlay $f_1$ with $d$ in position $p_1$\;
  resize $b_0$ with respect to $s_{b,0}$\;
  overlay $f_1$ with $b_0$ in position $p_{b,0}$\;

  draw $(p_2,s_2,f_2)$ in the same way\;
  draw a random bird $b_1$ from $B$\;
  draw $(p_{b,1},s_{b,1})$ for bird where $s_{b,1}$ is drawn from greater half of $S$\;
  resize $d$ with respect to $s_2$\;
  overlay $f_2$ with $d$ in position $p_2$\;
  resize $b_1$ with respect to $s_{b,1}$\;
  overlay $f_1$ with $b_1$ in position $p_{b,1}$\; \hspace*{5cm}
  
  save $f_0, f_1, f_2$ into the dataset
}

\end{algorithm}

\begin{table}[hbt!]
	\begin{center}
		\caption{Details of the dataset.
        \label{datasettable}}
		\vspace{-0.2cm}
		\begin{tabularx}{\columnwidth}{l l}
        	\hline
			\textbf{Aspect}  &  \textbf{Information}\\
			\hline \hline
			\# drones                            & 89\\
			\# birds                             & 126\\
			\# background videos                 & 11\\
			\# rows in grid                      & 12\\
			\# columns in grid                   & 10\\
			\# size intervals                    & 19\\
			size intervals                       & in [5,160] \\
            									 &(bias towards smaller values) \\
            image resolution                     & $850 \times 480$\\ 
			\hline
			\# resulting images                  & 676,534\\
            \hline
		\end{tabularx}
	\end{center}
\end{table}

\par 
The last things to mention about our approach are the training and prediction procedures. After creating the artificial dataset, we have applied a commonly used technique called fine-tuning. In this technique the network is first trained with a different and more general dataset for a similar problem. This provides us with better initial points than random for the parameters of the network. Then, training is continued with the actual dataset for the actual problem. This technique is useful especially when the training data is scarce. After training, there comes the prediction for unseen data. Since the network is trained with two classes, the bird detections are eliminated after getting all predicted bounding boxes from the last layer. Than a threshold, which can be determined according to accuracy on a validation set (our approach is explained in Section \ref{exp}), is applied on the confidence values for objectness. If this operations eliminate all predicted bounding boxes, it means that the frame does not include a drone or it is not clear enough to detect. Otherwise, the one that has the highest confidence is selected as the prediction. We choose the best prediction to report since the aforementioned challenge requires to detect the only drone in the scene. However, the algorithm can easily be extended to multi-drone situations with more intelligent thresholding strategies. One possible problem with this approach is encountered when the network mixes a bird up with the drone. If the objectness confidence of it is higher than that of the drone, it is selected as the prediction. In order to decrease the number of such misinterpretations, we propose a \textit{limited ignorance approach}. After determining the bounding box that the network is most confident, we control its intersection with the rectangle having same center, three times the width and height as the predicted bounding box in the previous frame, assuming that the drone cannot move more than its height or width in a single frame. If the rectangles intersect, we can accept the newly predicted one. Otherwise, we ignore the current prediction and report the previous one if the limit has not been exceeded yet. After exceeding the limit, we reset it and cancel the technique for the same number of frames. During this period, we report the current predictions directly. Likewise, when there is no predicted bounding box in the previous frame, we directly report the prediction in current frame.

\section{Experiments}
\label{exp}
This section describes training details and the conducted experiments on the artificial dataset and the real dataset provided by the organizers of the challenge. The former ones are evaluated quantitatively whereas the others are evaluated qualitatively due to the lack of ground truth information.
\par 

\textbf{Training details:} In order to apply fine tuning mentioned in Section \ref{method}, we have started with the pre-trained weights using the ImageNet dataset \cite{russakovsky2015imagenet} for image classification problem. Then the dataset provided by the challenge organizers and the created one are divided into training (85\%) and validation (15\%) parts. The training part of the former one is duplicated four times before combining them to training sets since it is too scarce compared to the artificially created, large scale one. Then, the network is fine-tuned for 10,000 iterations with 128 as batch size and batch normalization after all convolutional layers.
\par 
After the training phase is completed, we combined the two validation sets to evaluate the resulting network. Although we use $480 \times 480 \times 3$ as input size in training (see Figure \ref{yolomodel}), we increase the resolution to $800 \times 800 \times 3$ in testing configuration. This is applicable since the network is fully convolutional. This increase is helpful in detecting small sized targets.
\par

\textbf{Evaluation metrics:} We use precision-recall curves to evaluate the network. The curves are constructed by changing the detection threshold. The precision metric is defined as $\frac{tp}{tp+fp}$, where $tp$ is the number of true positives and $fp$ is the number of false positives. Recall is then defined as $\frac{tp}{tp+fn}$, where $fn$ is the number of false negatives. We count a predicted bounding box as true positive if the area of the overlap of the predicted bounding box with the ground truth is greater than half of the area of their union.

Another metric that we used is the prediction penalty, which is basically the area of smallest rectangle that includes both the ground truth and predicted bounding boxes divided by the area of ground truth bounding box.

\begin{figure}
\centering
	\includegraphics[width=\columnwidth]{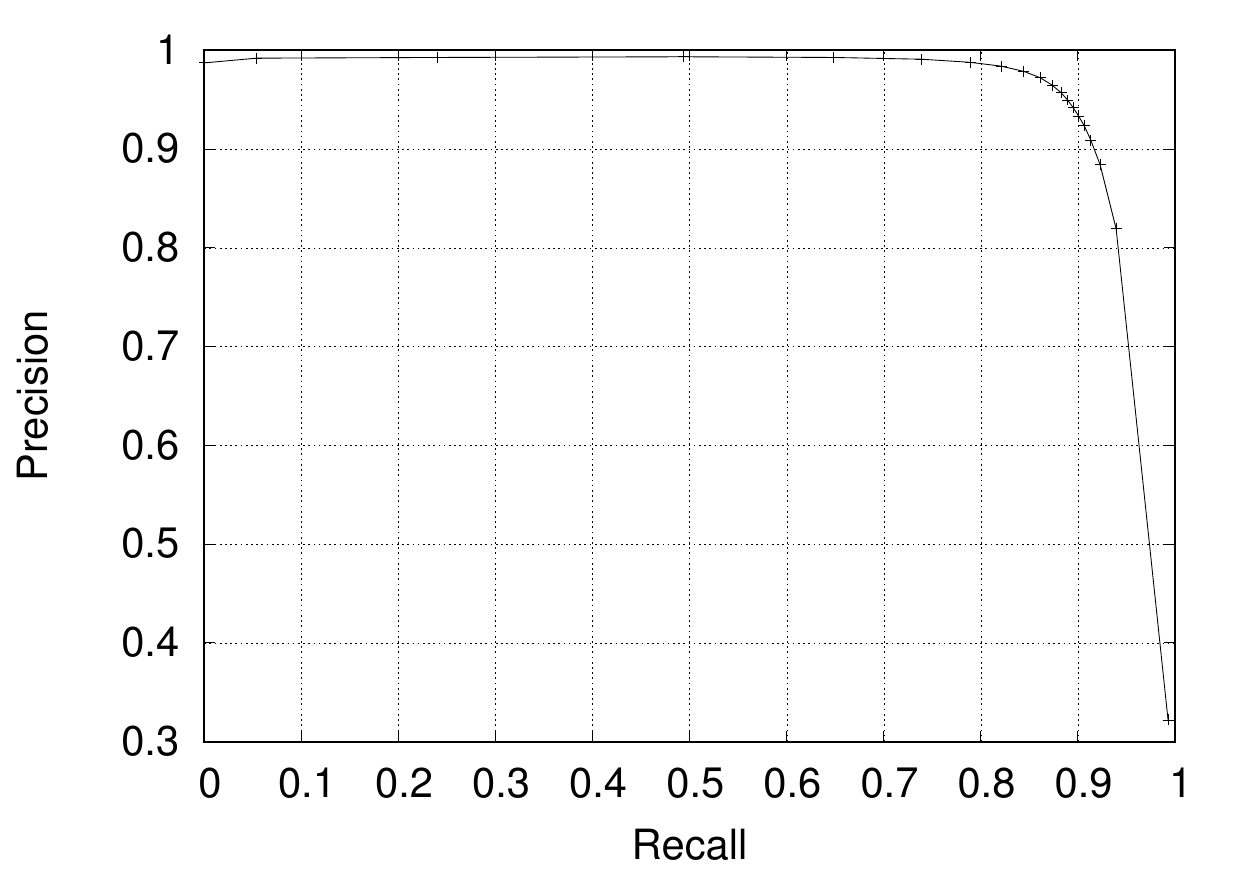}
    \caption{Precision-Recall (PR) curve showing the performance of the approach on the outdoor test videos.}
    \label{prcurve}
\end{figure}

\textbf{Results:} Figure \ref{prcurve} presents the performance of the method with different detection thresholds in the range [0,1]. The closer the Precision-Recall (PR) curve to the top right corner the better the performance of the method. We can understand from the curve that precision and recall can be achieved to be approximately 0.9 at the same time. This shows that the approach performs well in detecting the correct bounding boxes.
\par
\begin{figure}
\centering
	\includegraphics[width=\columnwidth]{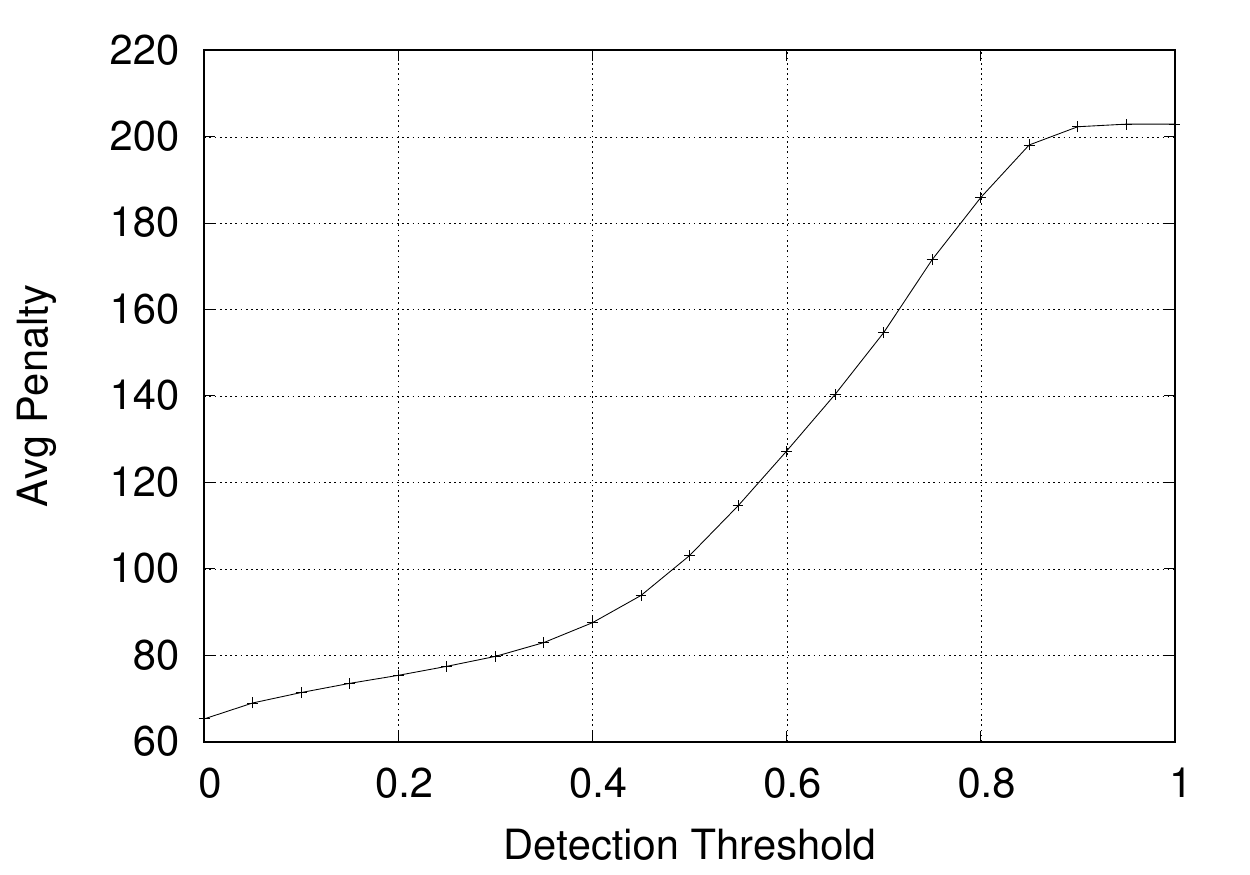}
    \caption{Change of prediction penalty with respect to detection threshold.}
    \label{penalty}
\end{figure}

Figure \ref{penalty} shows the change of the average penalty with respect to detection threshold. The reason for higher penalties is that when the threshold increases detection rate decreases.  When a drone cannot be found, the top-left pixel is reported as prediction, which results in a huge penalty. Hence, we have chosen the smallest possible threshold (which is zero) for quantitative evaluation on the test video of the challenge. Although this threshold hurts precision in the artificial dataset, it works well in the provided test video except its detecting the bird as drone when the bird is closer to the camera and in specific poses that cannot be easily distinguished from a drone by human eye. Another observation is that when the drone and the bird are too close to each other, the network supposes that the bird is a part of the drone, and outputs a bounding box enclosing both of them. The predictions are provided online\footnote{\url{http://user.ceng.metu.edu.tr/~cemal/predictions.mp4}} as a video rendered in 15 fps.

\section{Conclusion}
In this study, we showed that drones can be detected and distinguished from birds using an object detection model based on a CNN. The trained network generalizes well as it can achieve high precision and recall values at the same time. 
\par
For future work we plan to consider time domain to improve the performance even further. Since collecting such data is not easy, we plan to devise an algorithm that generates random flight videos instead of randomly generated images.

{\small

}


\end{document}